\documentclass[letterpaper]{article} 
\usepackage{aaai23}  
\usepackage{times}  
\usepackage{helvet}  
\usepackage{courier}  
\usepackage[hyphens]{url}  
\usepackage{graphicx} 
\usepackage{natbib}  
\usepackage{caption} 
\usepackage{algorithm}
\usepackage{algorithmic}
\usepackage{graphicx}
\usepackage{amsmath}
\usepackage{diagbox}
\usepackage{newfloat}
\usepackage{listings}
\usepackage{cite}
\usepackage{multirow}
\usepackage{url}
\usepackage{amsfonts,amssymb}

\usepackage[table,xcdraw,dvipsnames]{xcolor}

\usepackage[breaklinks,colorlinks,citecolor = ForestGreen]{hyperref}

\usepackage{booktabs}
\usepackage{hyperref}

\DeclareCaptionStyle{ruled}{labelfont=normalfont,labelsep=colon,strut=off} 
\floatstyle{ruled}
\newfloat{listing}{tb}{lst}{}
\floatname{listing}{Listing}
\title{Learning Motion-Robust Remote Photoplethysmography\\through Arbitrary Resolution Videos}
\author{
    Jianwei Li\textsuperscript{\rm 1*},
    Zitong Yu\textsuperscript{\rm 2}\thanks{Equal contribution},
    Jingang Shi\textsuperscript{\rm 1}\thanks{Corresponding author}
}
\affiliations{
    \textsuperscript{\rm 1}School of Software Engineering,Xi’an Jiaotong University\\
    \textsuperscript{\rm 2}School of Electrical and Electronic Engineering,Nanyang Technological University\\

    3121158008@stu.xjtu.edu.cn, zitong.yu@ieee.org, jingang@xjtu.edu.cn
}

\usepackage{bibentry}

\begin{document}

\maketitle

\begin{abstract}
Remote photoplethysmography (rPPG) enables non-contact heart rate (HR) estimation from facial videos which gives significant convenience compared with traditional contact-based measurements. In the real-world long-term health monitoring scenario, the distance of the participants and their head movements usually vary by time, resulting in the inaccurate rPPG measurement due to the varying face resolution and complex motion artifacts. Different from the previous rPPG models designed for a constant distance between camera and participants, in this paper, we propose two plug-and-play blocks (i.e., physiological signal feature extraction block (PFE) and temporal face alignment block (TFA)) to alleviate the degradation of changing distance and head motion.
On one side, guided with representative-area information, PFE adaptively encodes the arbitrary resolution facial frames to the fixed-resolution facial structure features. On the other side, leveraging the estimated optical flow, TFA is able to counteract the rPPG signal confusion caused by the head movement thus benefit the motion-robust rPPG signal recovery. Besides, we also train the model with a cross-resolution constraint using a two-stream dual-resolution framework, which further helps PFE learn resolution-robust facial rPPG features. Extensive experiments on three benchmark datasets (UBFC-rPPG, COHFACE and PURE) demonstrate the superior performance of the proposed method. One highlight is that with PFE and TFA, the off-the-shelf spatio-temporal rPPG models can predict more robust rPPG signals under both varying face resolution and severe head movement scenarios. The codes are available at \href{https://github.com/LJW-GIT/Arbitrary\_Resolution\_rPPG}{https://github.com/LJW-GIT/Arbitrary\_Resolution\_rPPG}.

\end{abstract}

\section{Introduction}

Heart rate (HR) is an important physiological signal which is widely used in many circumstances, especially for healthcare or medical purposes. Electrocardiography (ECG) and Photoplethysmograph (PPG)/Blood Volume Pulse (BVP) are the two most common methods of measuring heart activities. However, these sensors need to be attached to body parts, limiting their usefulness and scalability. Due to the inconvenience of long-term monitoring and discomfort for the users, traditional ways limit the application scenarios such as driver status assessment and burn patient health monitoring. To solve this problem, non-contact HR measurement, which aims to measure heart activity remotely, has become an increasingly popular research problem in physiological signal measurement in recent years.

Most existing non-contact HR measurement approaches are based on the facial video-based remote Photoplethysmograph (rPPG) technique~\cite{yu2021facial}. The rPPG method uses digital cameras to record variations of reflected ambient light on facial skin, which contains information on cardiovascular blood volume and pulsation. However, the rPPG measurement is very susceptible and vulnerable to the quality of video recording and head motions. In the early stage, handcrafted features based methods~\cite{takano2007heart, verkruysse_remote_2008} require an exhausted multi-stage process (preprocessing, filtering and post-processing) and are with low robustness to head motions and illumination changes. Thus, they are usually tested and deployed under controlled lab environment scenarios.

\begin{figure}[t]
\centering
\includegraphics[width=0.9\columnwidth]{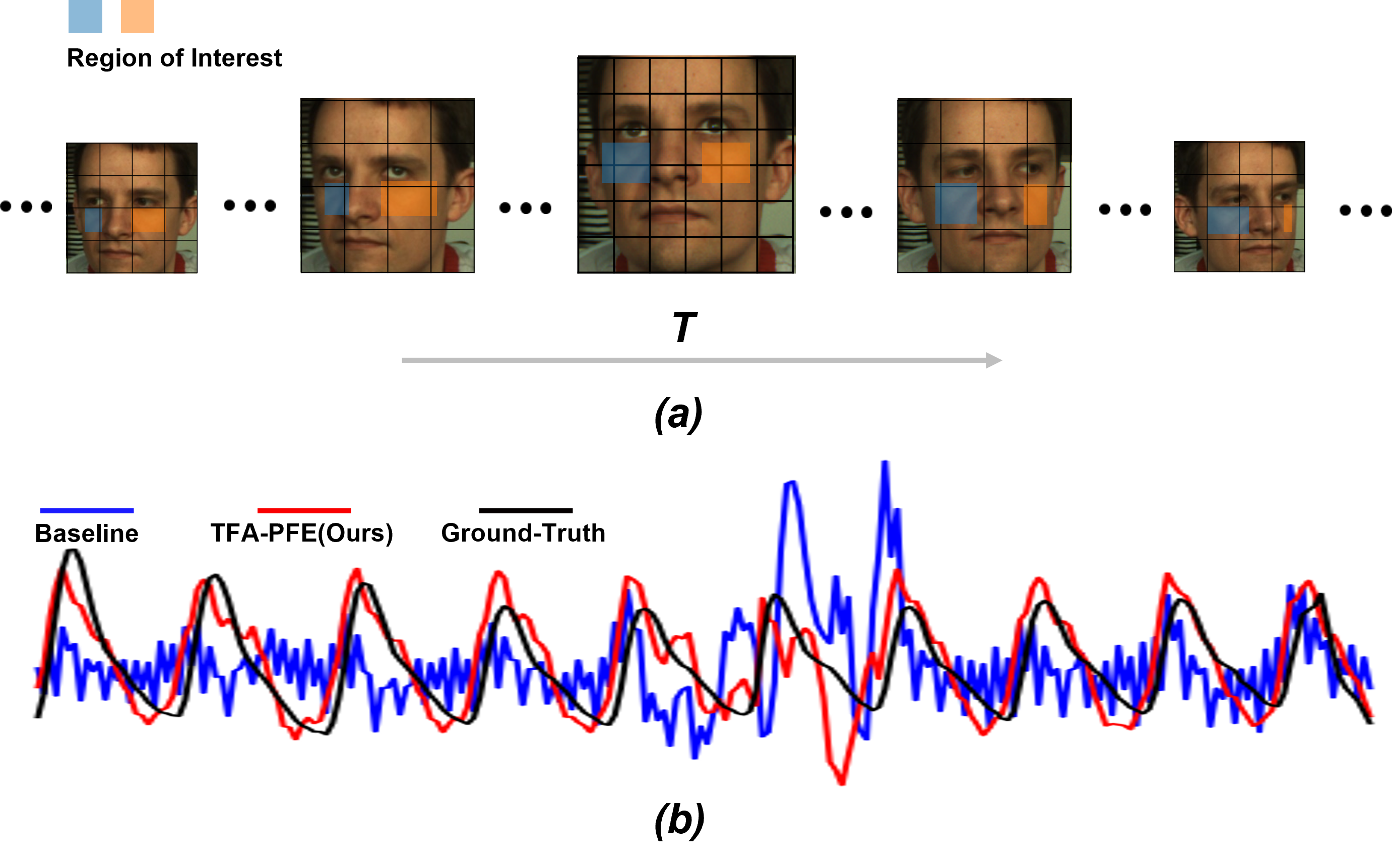} 
\vspace{-1.0em}
\caption{rPPG measurement from arbitrary resolution videos with head movements. (a) The ROIs might have different spatial size and shape across the temporal dimension. (b) Compared with the baseline PhysNet~\cite{yu_remote_2019}, the proposed TFA-PFE can predict more accurate rPPG signals with the ground truth BVP signals.}
\label{fig1}
\vspace{-0.8em}
\end{figure}

With the rapid development of deep learning, neural network models have also been widely applied in the rPPG field. Recent spatio-temporal representation map~\cite{niu2020video, lu_dual-gan_2021} based and end-to-end~\cite{yu_remote_2019,chen2018deepphys} based models utilize Convolutional Neural Networks (CNNs) to learn the spatio-temporal rPPG cues from facial videos with fixed resolution, which have shown superior performance compared with traditional approaches. However, most of the existing rPPG approaches rarely consider the real-world practical situation with arbitrary resolution face videos (e.g., the distance of the participants varies by time, see Figure~\ref{fig1}(a) for visualization). Previous methods~\cite{mcduff2018deep} simply use the spatial interpolation method to scale the face frames with arbitrary resolutions to a fixed size to adapt the model input, where the vital pixels of the region of interest (ROI) might be confused by the interpolation method, thus harming the accuracy of rPPG measurement. 
Meta-SR ~\cite{hu_meta-sr_2019} and LIIF ~\cite{chen_learning_2021} bring the scale-arbitrary super-resolution problem into the horizon.
To our best knowledge, there is still no solution proposed yet to counter the problem of rPPG measurement from arbitrary face resolution videos.

Besides arbitrary face resolution, another noteworthy issue of rPPG measurement is the motion-robustness. Due to the limited spatio-temporal receptive field of CNN with weak capacity of spatial contextual ROI localization, both rigid and non-rigid head motions usually have serious impacts on rPPG measurement. Existing end-to-end models~\cite{yu_remote_2019,yu2021transrppg} do not take initiative (e.g., face alignment operation) to describe the head movement to compensate the motion artifacts, resulting the vulnerability to severe head movement. Few works passively adopt the ROI tracking~\cite{niu_vipl-hr_2018} or utilize normalized frame difference motion representation as input~\cite{chen2018deepphys,liu_multi-task_2020}. However, these methods have limited stability and are hard to directly plug in the off-the-shelf end-to-end spatio-temporal rPPG models.

Motivated by the discussions above, we propose two plug-and-play blocks (i.e., Physiological Signal Feature Extraction block (PFE) and Temporal Face Alignment block (TFA)) to capture resolution- and motion-robust rPPG features. To learn the similarity of rPPG signals from arbitrary resolution frames, we design a new cross-resolution constraint using a dual-resolution framework, which further helps PFE learn resolution-robust facial rPPG features. As shown in Figure~\ref{fig1}(b), compared with the vanilla PhysNet~\cite{yu_remote_2019}, the proposed TFA and PFE blocks can benefit more accurate rPPG measurement from arbitrary resolution videos with head movements. The contributions of this work are as follows:
\begin{itemize}
    \item To our best knowledge, we provide the first solution to plug-and-play modules on robust rPPG measurement from facial videos with arbitrary fixed resolution or varying face resolution. 
    
    \item We propose the PFE block to adaptively encode the arbitrary resolution facial frames to the fixed-resolution facial structure features. Besides, we propose to train the model with a cross-resolution constraint using a two-stream dual-resolution framework, which further helps PFE learn resolution-robust facial rPPG features.
    
    \item We propose the TFA block to counteract the rPPG signal confusion caused by the head movement via wrapping facial frames by the estimated optical flow, which benefits the motion-robust rPPG signal recovery and alleviates the influence of head movement.
    
    \item We conduct extensive experiments on benchmark datasets to demonstrate the superior performance of the proposed method under both arbitrary face resolution and severe head movement scenarios.
\end{itemize}

\section{Related Work}
\subsection{Remote Photoplethysmography Measurement}
Plenty of handcrafted rPPG measurement methods have been proposed since the researches~\cite{takano2007heart, verkruysse_remote_2008} show the feasibility of recovering physiological signals through a digital camera. Some early works take traditional signal processing methods into consideration, which contain matrix transformation ~\cite{tulyakov_self-adaptive_2016,shi2019atrial}, Least Mean Squares ~\cite{li2014remote}, and Blind Source Separation (BSS) ~\cite{poh2010advancements, poh2010non}. In recent years, with the boost of deep learning methods, Deephys ~\cite{chen2018deepphys} and PhysNet~\cite{yu_remote_2019} firstly introduce end-to-end based CNN framework to this field. Meanwhile, spatio-temporal signal map based methods~\cite{niu2020video,lu_dual-gan_2021} also attract more attention due to their excellent performance. Towards efficient rPPG measurement, Auto-HR \cite{yu_autohr_2020} and EfficientPhys ~\cite{liu_efficientphys_2022} search and design lightweight end-to-end models. Recently, PhysFormer \cite{yu_physformer_2022} gains progress via temporal difference transformers to explore the long-range spatio-temporal relationship for rPPG representation. Besides supervised learning with labeled facial videos, unsupervised learning has also been validated in ~\cite{gideon_way_2021} to achieve rPPG measurement. The vulnerability of rPPG models has also been discussed recently, such as phase difference~\cite{mironenko_remote_2020, moco_impairing_2018}, camera rolling ~\cite{moco_impairing_2018, zhan_analysis_2020} and video compression format ~\cite{mcduff_impact_2017}.


\begin{figure*}[t]
\centering
\includegraphics[width=0.75\textwidth]{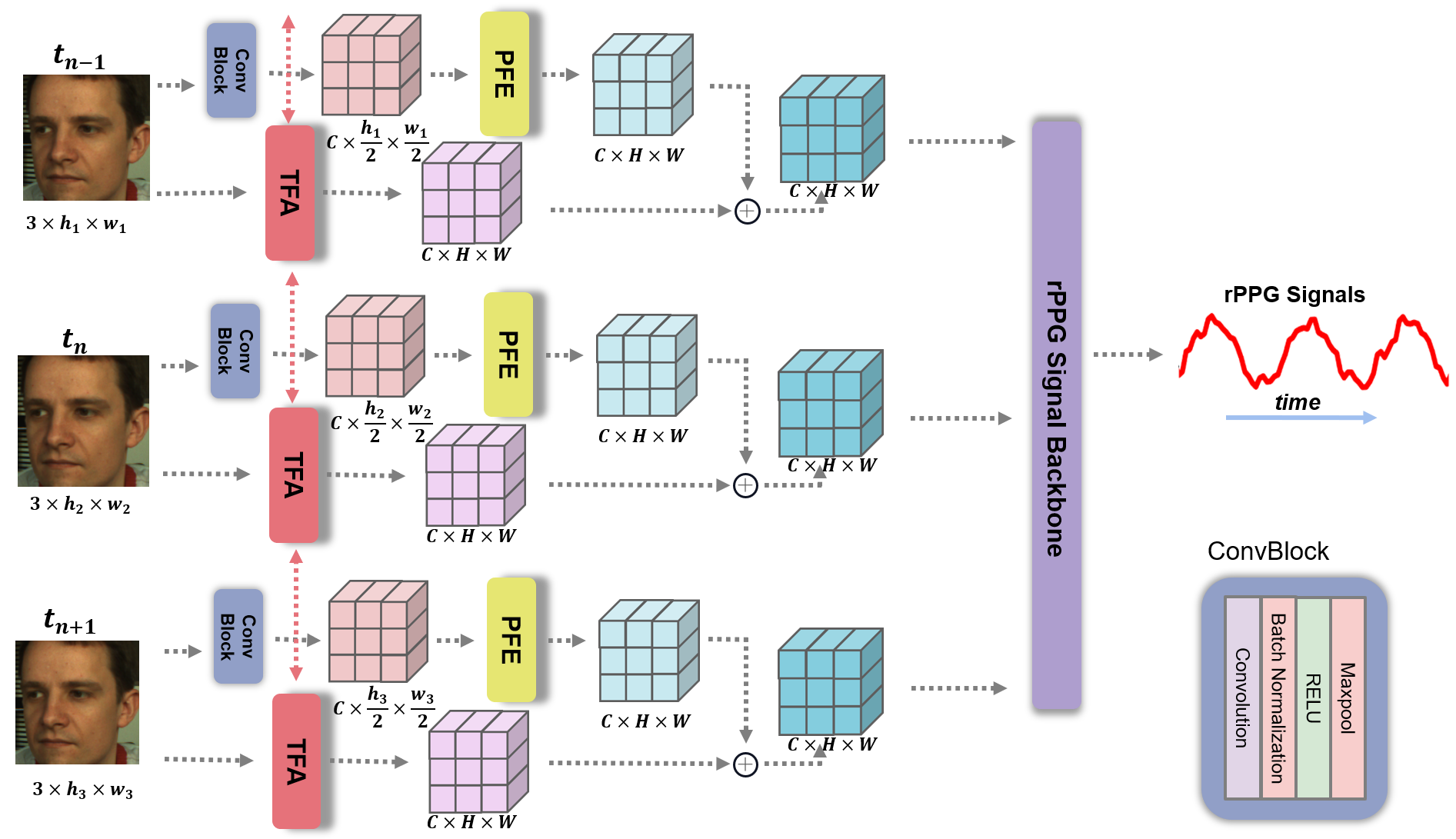} 
\vspace{-0.8em}
\caption{Overall framework of the proposed method. The frames in the sequence can be the arbitrary resolution. In the spatial stream, each arbitrary resolution face frame in the sequence forwards the Physiological Signal Feature Extraction (PFE) block, which maps the arbitrary size features to the fixed size facial structure features. In the temporal stream, the Temporal Face Alignment (TFA) block interpolates the frames to the same shape to generate the temporal aligned features. The facial structure and temporal aligned features are added to form the facial structure-motion features. Finally the facial structure-motion features forwards a rPPG Signal backbone to predict the rPPG signals.}
\label{fig2}
\end{figure*}

\subsection{Face Resolution and its Impact for RPPG}

In real-world applications, the distances between the camera and participants are variant, resulting in arbitrary resolution on facial region. Some recent works aim to extract rPPG from low-resolution videos with fixed size. They use super-resolution models through the end-to-end ~\cite{mcduff2018deep} or two-stage~\cite{song2020new,yue2021deep} network to recover the high-quality face videos and corresponding rPPG signals simultaneously. However, the target of super-resolution~\cite{shi2018hallucinating,shi2019face} is mainly toward the recovery of visual performance, but not to maintain the quality of rPPG signals. Furthermore, it is still a challenge to obtain reasonable rPPG signals in the arbitrary face resolution scenario, which is more practical in the real-world application. 

\subsection{Face Alignment for RPPG}
The alignment of face video enhances the performance of rPPG measurement. One straightforward evidence is that with the assistance of face landmark ~\cite{xia_sparse_2022, wan2020robust, shiidpt}, generated facial ROI-based spatio-temporal signal maps~\cite{lu_dual-gan_2021, lu2021hr, niu2018synrhythm, niu2020video, niu_vipl-hr_2018} benefit the motion-robust rPPG measurement. However, the efficacy of these methods highly depend on the accuracy of the detected face landmarks. As reported in~\cite{niu_vipl-hr_2018}, the huge rotation of head easily causes the loss of landmarks. 
Recently, temporal alignment is wildly applied in video frame interpolation and video super-resolution. Optical flow calculated for image-level alignment ~\cite{yi2019progressive} and feature-level alignment ~\cite{chan_basicvsr_2021} are two representative streams for temporal alignment. In this work, we introduce the optical flow guided feature alignment for motion-robust rPPG measurement. 

\vspace{-0.5em}
\section{Methodology}
\subsection{Overall Framework} 
As illustrated in Figure~\ref{fig2}, given a arbitrary resolution face sequence ${ X = \left [  x_{1} , x_{2},...,x_{i}  \right ] , x_{i} \in \mathbb{R}^{3\times h_{i}\times w_{i}}}$, ${  i\in \left\{ 1,...,T\right\}}$ as input, the proposed method forwards the two-stream pathway including the Physiological Signal Feature Extraction block (PFE) and the Temporal Face Alignment block (TFA) to form the facial structure-motion features. Then the rPPG backbone (e.g., PhysNet~\cite{yu_remote_2019}) is used for rPPG signals prediction. Please note that the arbitrary width ${h_{i}}$ and height ${w_{i}}$ of each frame can be different.

Before the PFE stream, we adopt a ConvBlock to extract features ${X_{ar} = \left [ x_{ar}^{1}, x_{ar}^{2},...,x_{ar}^{i} \right ]}$, ${x_{ar}\in \mathbb{R}^{C\times \frac{h_{i}}{2} \times \frac{w_{i}}{2}}}$ from the arbitrary resolution face sequence ${ X }$. Specifically, the ConvBlock is formed by a convolutional block with kernel size (${1\times 5\times 5}$) cascaded with batch normalization (${BN}$), ${RELU}$, and ${MaxPool}$, where the pooling layer halves the spatial dimension. Then ${X_{ar}}$ forwards the PFE block to generate facial structure features ${X_{st}\in \mathbb{R}^{T\times C\times H \times W}}$. ${T,C,H,W}$ indicate constant clip length, channel, height and width, respectively.

In term of the TFA stream, the TFA block first interpolates the arbitrary resolution sequence $X$ to ${\hat{X}\in \mathbb{R}^{T\times3\times H\times W} }$, which has the same height and width as ${X_{st}}$. Then, TFA uses the bi-directional optical flow (forward and backward) from successive frames to obtain temporal face alignment features ${X_{mo} \in \mathbb{R}^{T\times C \times H\times W}}$.

The output of PFE ${X_{st}}$ and the output of TFA ${X_{mo}}$ are summed to form the facial structure-motion features ${X_{st-mo} \in \mathbb{R}^{T\times C \times H\times W} }$. Finally, the rPPG signal backbone predicts the 1D rPPG signals ${Y \in \mathbb{R}^{T}}$ from ${X_{st-mo}}$. More details for reproducibility could be found in our code.


\subsection{Physiological Signal Feature Extraction (PFE)}

\begin{figure}[t]
\centering
\includegraphics[width=1.0\columnwidth]{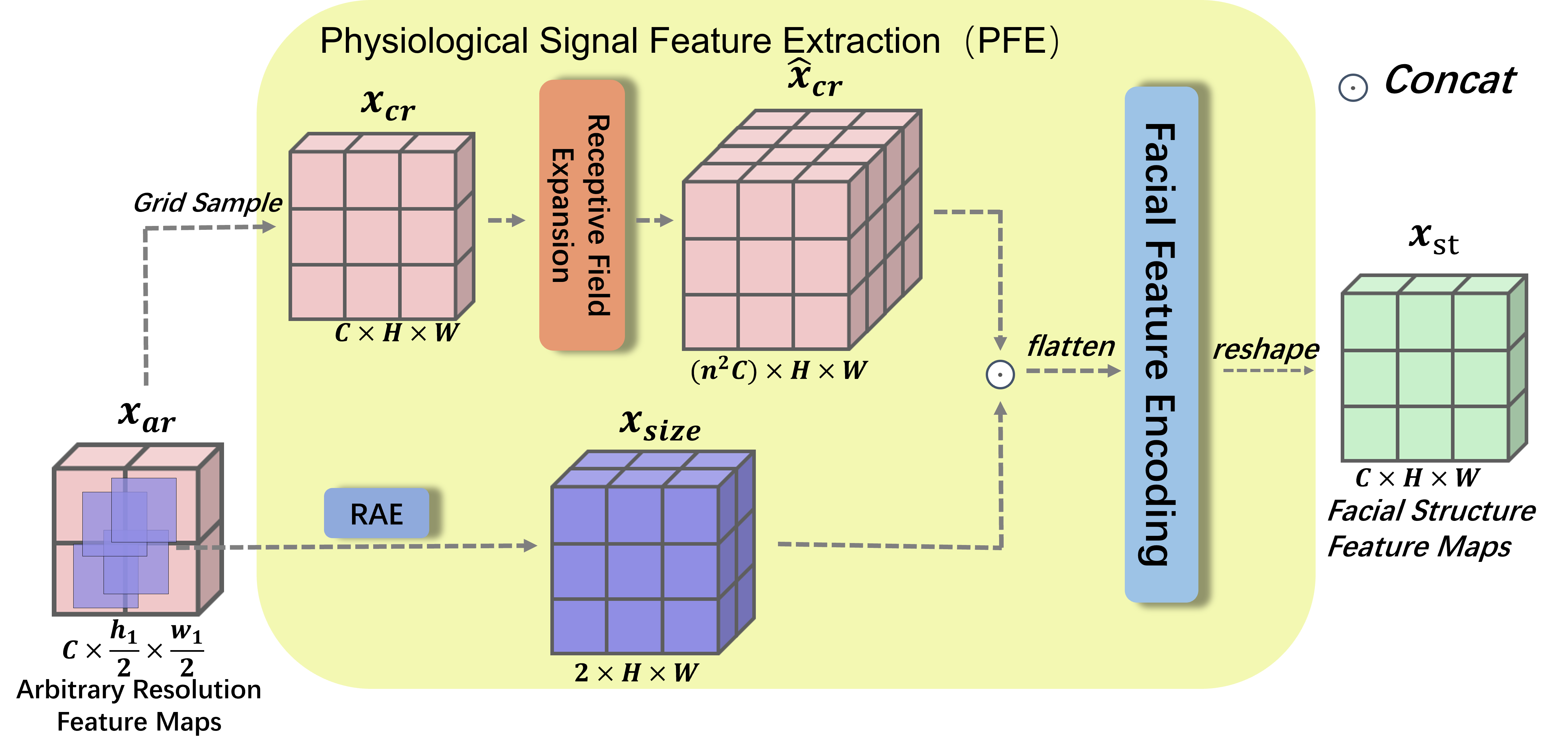} 
\vspace{-1.8em}
\caption{The structure of the PFE block.}
\label{pfe}
\vspace{-0.8em}
\end{figure}

The PFE block is used in the spatial dimension for each face frame. As shown in Figure~\ref{pfe}, the proposed PFE contains two parts. The upper and lower branches are devised for facial information and position information respectively. For upper branch, the features ${x_{ar}}$ from arbitrary-resolution frame are first interpolated to constant resolution features ${{x}_{cr} \in  \mathbb{R}^{C\times  H \times W}}$. To exploit the facial features, \textbf{receptive field expansion} is conducted as Eq.(\ref{equation2}) to obtain the expanded features ${\hat{x}_{cr} \in \mathbb{R}^{({n^{2}C})\times  H \times W} }$. Meanwhile, in the lower branch, \textbf{Representative area encoding (RAE)} is employed as Eq.(\ref{equation3}) to record the mapping relationship of pixel positions between ${x_{ar}}$ and ${x_{cr}}$. The relationship is described as coordinate tensor ${x_{size} \in \mathbb{R}^{{2}\times  H \times W}}$, where two channels represent the scaling ratio on the height and width accordingly. Then, the expanded features ${\hat{x}_{cr}}$ together with coordinate tensor ${x_{size}}$ are fed into the \textbf{facial feature encoding} as Eq.(\ref{equation1}) to produce the facial structure features ${x_{st} \in  \mathbb{R}^{C\times  H \times W}}$.

\subsubsection{Receptive field expansion.}
To enrich and mine the contextual information contained in  the facial structure features ${x_{cr}}$, we unfold the facial structure features ${x_{cr}}$ first, and then expand its receptive field via concatenating the ${n\times n}$ neighboring features to obtain ${\hat{x}_{cr} }$. Formally, the receptive field expansion is defined as \begin{equation}
\hat{x}_{cr}(i,j)  = Concat(\left \{ x_{cr}(i+n,j+n)\right \} _ {n\in Neighbor}), 
\label{equation2}
\end{equation}
where $i$ and $j$ indicate the spatial position of the features. $n=3$ is used as the default setting. 


\subsubsection{Representative area encoding (RAE).}
As the arbitrary size features ${x_{ar}}$ have different spatial size with the structure features ${x_{st}}$, spatial positions in ${x_{st}}$ correspond to different areas from ${x_{ar}}$. It is important to explicitly describe the representative area for each spatial position. Here we formulate the representative area information ${x_{size} \in \mathbb{R}^{2\times H\times W}}$ as 
\begin{equation}
\begin{split}
{x_{size}}\left ( i,j \right )  = [\sigma _{H},\sigma _{W}], \\
\sigma _{H} = \frac{h}{H}, \sigma _{W} = \frac{w}{W},
\end{split}
\label{equation3}
\end{equation}
where ${\sigma _{H}}$ and ${\sigma _{W}}$ mean the scaling ratio on the height and width dimensions when ${x_{ar}}$ is transformed to ${x_{st}}$.

\subsubsection{Facial feature encoding.}
We concatenate ${\hat{x}_{cr}}$ and ${x_{size}}$ along the channel dimension. A shallow facial feature encoding function is designed to mine the semantic facial structure features, which is simply parameterized as an MLP. Facial feature encoding takes the form:
\begin{equation}
{x_{st}}= Reshape \left( MLP\left (Flatten( Concat \left( \hat{x}_{cr}, x_{size} ) \right) \right ) \right)
\label{equation1}
\end{equation}
After extracting the facial structure features ${x_{st} \in \mathbb{R}^{C\times H\times W}}$ from each frame, we merge them in temporal dimension to form the ${X_{st} \in \mathbb{R}^{T \times C\times H\times W}}$.

\subsection{Temporal Face Alignment (TFA)}


The facial structure features ${X_{st}}$ from the PFE block have rich representation capacity on the arbitrary resolution condition. However, in practice, head movement influences end-to-end rPPG measurement significantly. For example, huge-angle rotation will make partial facial features out of the scope of the facial structure features ${X_{st}}$. Previous works use landmark detection methods such as OpenFace~\cite{baltruvsaitis2016openface} to extract the face landmarks for facial ROI alignment. However, the robustness of rPPG measurement is highly depended on the accuracy of face landmarks. Here, three problems are noted for face alignment:
\begin{enumerate}
\item Landmarks status. The position of face landmarks could change dramatically because of head motion, which induces the inaccurate detection of ROIs in the facial clips.
\item Interpolation. The shape of ROI might be different, and interpolation is usually used to keep the consistency \cite{hu2021end}. However, interpolation may corrupt the color change of pixels, and eliminate the rPPG cues.
\item Lost landmarks. When the head movement encounters huge-angle rotation, partial face may disappear from the frame. In this case, the predicted landmarks would mark some regions randomly.
\end{enumerate}

\begin{figure}[t]
\centering
\includegraphics[width=1\columnwidth]{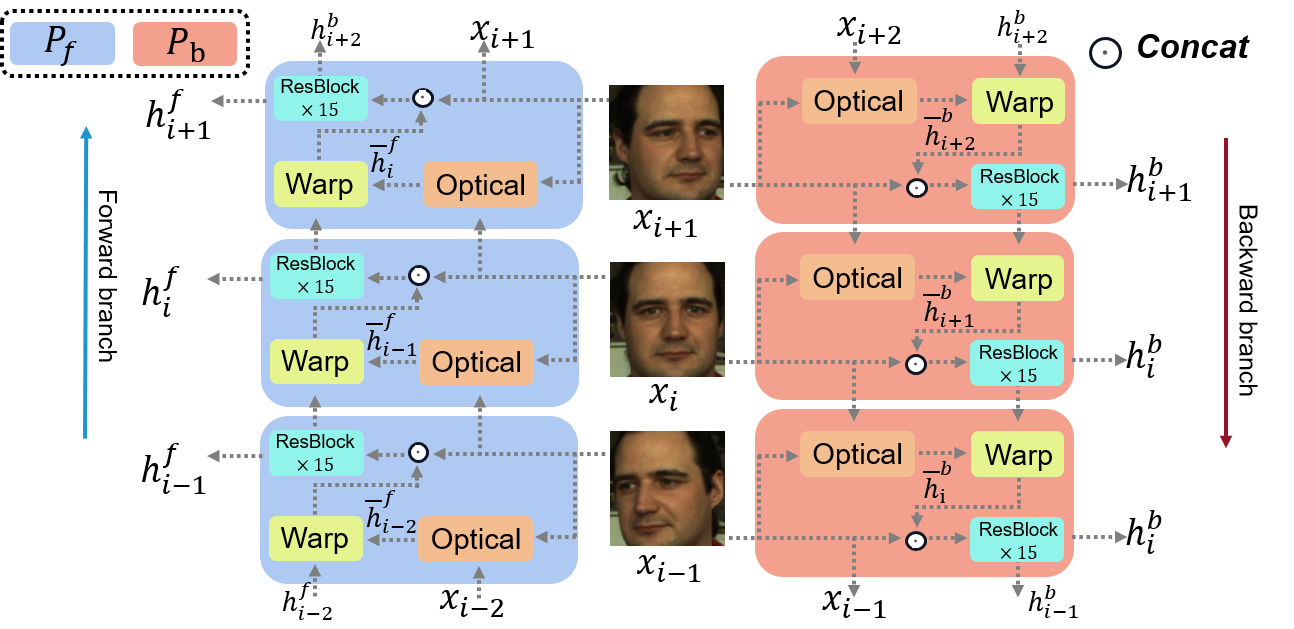} 
\vspace{-1.3em}
\caption{The structure of TFA block. ${P_{f}}$ and ${P_{b}}$ mean the forward and backward optical flow face alignment block, respectively. }
\label{fig5}
\vspace{-0.6em}
\end{figure}

Head rotation is a continuous process, and the state of each frame is correlated with the forward and backward states. According to these observations, we propose the temporal face alignment (TFA) block to leverage optical flow to describe the facial motion and wrap the facial structure features.

As shown in Figure \ref{fig5}, TFA adopts a typical bidirectional recurrent network. The video sequence ${\hat{X}}$ forwards \textbf{optical flow face alignment} as Eq.(\ref{equation5}) to get head motion features ${H^{b,f} = \left [ h_{1}^{b,f}, h_{2}^{b,f},..., h_{i}^{b,f} \right ]}$, ${h_{i}^{b,f}\in \mathbb{R}^{C\times H \times W}}$. Then, ${h^{b}}$ and ${h^{f}}$ are fed into \textbf{bidirectional aggregation} as Eq.(\ref{equation6}) to produce the facial structure features ${x_{st}}$.

\subsubsection{Optical flow face alignment.}

The video sequence ${\hat{X}}$ is first fed into ‘Optical’ to calculate the optical flow ${s_{i}}$ by SPyNet \cite{8099774}. Then, the optical flow ${s_{i}}$ is utilized to 'Warp' the head motion features ${{h}_{i-1}}$ of previous frame to get aligned features ${\bar{h}_{i-1}}$ for frontalizing faces. Notice status ${{h}_{0}}$ is initialized by features of all zeros. The aligned features together with the recent frame are then passed to 15 basic residual blocks for obtaining ${{h}_{i}}$. Optical flow face alignment takes the form:
\begin{equation}
\begin{aligned}
s_{i}^{b,f} &  = Optical\left( x_{i}, x_{i\pm 1} \right),  \\
\bar{h}_{i\pm 1} & = Wrap \left (h_{i\pm 1},s_{i}^{b,f} \right ),  \\
h_{i}^{b,f} & = ResBlock\left ( Concat \left ( x_{i}, \bar{h}_{i \pm 1} \right)  \right ), 
\label{equation5}
\end{aligned}
\end{equation}
The head motion features ${h_{i}}$ can be represented from two temporal directions (i.e., forward and backward). Thus, we use ${h_{i}^{f}}$ and ${h_{i}^{b}}$ to represent the ${h_{i}}$ via forwarding and backwarding ${x_{i}}$.


\subsubsection{Bidirectional aggregation.}
To aggregate the backward and forward head motion features, we concatenate  ${h_{i}^{b}}$ and ${h_{i}^{f}}$ along the channel dimension and introduce a convolutional layer to maintain the number of channels. Formally, the bidirectional aggregation is defined as
\begin{equation}
\begin{aligned}
x_{mo} &= F \left ( Concat \left (  {h_{i}^{b}, h_{i}^{f} } \right ) \right)
\end{aligned}
\label{equation6}
\end{equation}
where ${F \left( \cdot \right)}$ represents a ${1 \times 1}$ convolutional layer and ${x_{mo} \in \mathbb{R}^{C\times H\times W}}$ represents generated temporal face alignment features.

Finally, the facial structure features ${x_{st}}$ are added to ${x_{mo}}$ to obtain the facial structure-motion features ${{x}_{st-mo} \in \mathbb{R}^{C\times H\times W}}$.

\subsection{Cross-Resolution Constraint and Loss Functions}

\begin{figure}[t]
\centering
\includegraphics[width=0.9\columnwidth]{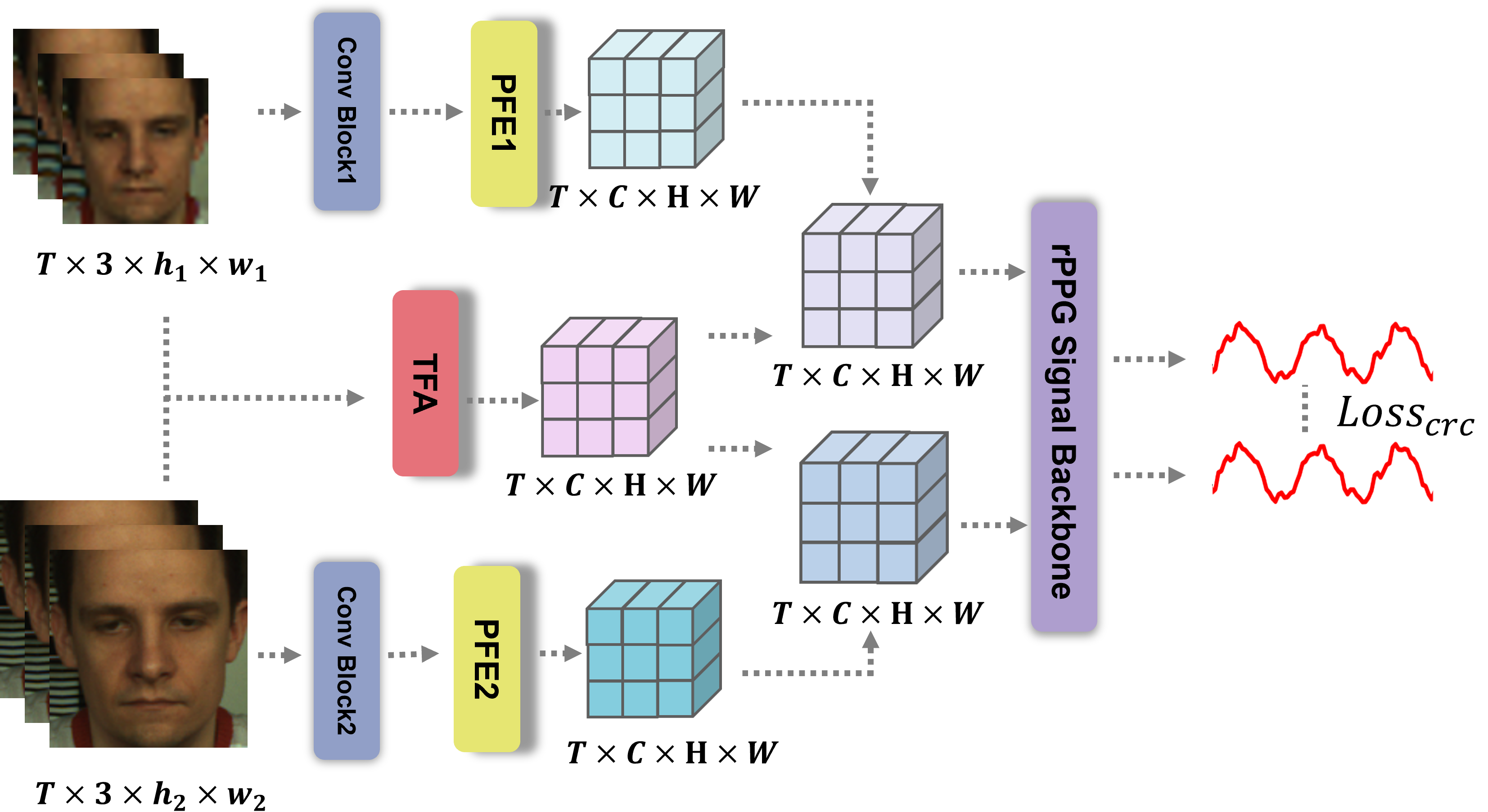} 
\vspace{-0.6em}
\caption{The framework of the Cross-Resolution Constraint. It is calculated on the two sequence inputs of arbitrary face images from the same clip.}
\label{doubel-branch}
\end{figure}

Despite we have designed the PFE block to tackle with arbitrary resolution problem, it is still hard to learn resolution-invariant rPPG features with the traditional Negative Pearson loss $\mathcal{L}_{time}$~\cite{yu_remote_2019} and frequency cross-entropy loss $\mathcal{L}_{fre}$~\cite{niu2020video}. Here we design a novel cross-resolution constraint $\mathcal{L}_{crc}$ which forces the model to learn consistent rPPG predictions between two resolution views. Specifically, as shown in Figure~\ref{doubel-branch}, we sample video clip into different resolutions as  $X_{1}\in \mathbb{R}^{T\times3\times h1\times w1}$ and $X_{2}\in \mathbb{R}^{T\times3\times h2\times w2}$. The two sampled clips forward the unshared PFE and shared TFA blocks first, and then go through a shared rPPG backbone model to predict the corresponding rPPG signals $Y_{1}\in \mathbb{R}^{T\times1}$ and $Y_{2}\in \mathbb{R}^{T\times1}$. The cross-resolution constraint $\mathcal{L}_{crc}$ can be formulated via calculating the L1 distance between two predicted signals. The overall loss function $\mathcal{L}_{overall}$ can be formulated as
\begin{equation}
\begin{split}
\mathcal{L}_{crc}&=\left. \| Y_{1}-Y_{2} \right. \| _{1},\\
\mathcal{L}_{overall}&=\mathcal{L}_{time}+\mathcal{L}_{fre}+\alpha\cdot\mathcal{L}_{crc},
\end{split}
\end{equation}
where hyperparameter $\alpha$ equals to 0.1. The loss function avoids the model to pay attention to the similarity of low-level features from different resolution. In other words, $\mathcal{L}_{crc}$ focuses on the consistency of predicted rPPG signals, instead of the feature-level consistency, which determines the performance measurement and provide the direct supervision signals for the model learning.

\section{Experiment}

We first conduct experiments of rPPG-based HR measurement on three benchmark datasets with their original protocols and normal setting. Then, the UBFC-rPPG~\cite{bobbia2019unsupervised} dataset is used for performance evaluation on arbitrary-resolution facial videos and ablation studies.

\subsection{Dataset}

\subsubsection{UBFC-rPPG.}
The UBFC-rPPG dataset~\cite{bobbia2019unsupervised} includes 42 videos, which are about 2 minutes long and recorded. The bio-signals ground-truth were recorded by a pulse oximeter with a 60 Hz sampling rate.

\subsubsection{PURE.}
The PURE dataset~\cite{stricker2014non} contains 60 videos from 10 subjects performing six different head motion tasks: steady, talking, slow translation, fast translation, small rotation, and medium rotation.

\subsubsection{COHFACE.}
The COHFACE dataset~\cite{heusch2017reproducible} is consisted of 160 one-minute videos from 40 healthy individuals, captured under studio and natural light. The videos are heavily compressed using MPEG-4 Visual, which was noted by \cite{mcduff_impact_2017} to potentially cause corruption of the rPPG signal. 


\subsection{Implementation Details}
\subsubsection{Preprocessing and training procedure.}
For each video clip, we use the MTCNN \cite{zhang2016joint} to crop the enlarged face area and resize each frame to ${128\times128}$ pixels. And then we downsample the face image ranging from 1.0 to 4.0 times to get the arbitrary scale frames. The facial video clip with arbitrary sizes would be mapped to the fixed size $H$=$W$=64 after PFE and TFA blocks while $C$=16. Random horizontal flipping is used for clip-level spatial data augmentation. The proposed method is trained with batchsize 2 on RTX3090 GPU with PyTorch. The Adam optimizer is used and the learning rate is set as 1e-4. The weight decay is 5e-5.


\subsubsection{Metrics and evaluation.}
Following \cite{comas_efficient_2022}, we calculate the root mean squared error (RMSE) and mean absolute error (MAE) between the predicted average HR versus the groundtruth HR. We first forward the models using 160-frame clips without overlapping to predict clip-level HR. To fairly compare our method with the state-of-the-art methods~\cite{vspetlik2018visual,comas_efficient_2022,lokendra2022and}, the whole-video performance comparisons are calculated via averaging the clip-level predictions.


\subsection{Comparison on Normal Face Resolution Setting}

For fair comparison, we train the baseline model PhysNet~\cite{yu_remote_2019} and our methods using the same recipe to alleviate the influence of arbitrary resolution and data resolution augmentation. As shown in Table \ref{tabel2}, the performance of our re-implemented PhysNet is slightly better than the result reported in ~\cite{gideon_way_2021}. The UBFC-rPPG (UBFC) dataset is used for the ablation study about the proposed TFA and PFE blocks under  arbitrary-resolution and head motion scenarios. We compare our method with eight state-of-the-art methods (CHROM~\cite{de2013robust}, POS~\cite{wang2016algorithmic}, HR-CNN~\cite{vspetlik2018visual}, DeepPhys~\cite{chen2018deepphys},Zhan et al.~\cite{zhan_analysis_2020}, Gideon et al.~\cite{gideon_way_2021}, AND-rPPG~\cite{lokendra2022and}, TDM~\cite{comas_efficient_2022}) in Table \ref{tabel1}.


\begin{table}[t]
\centering
\caption{HR estimation results (bpm) on UBFC, PURE, and COHFACE datasets. The proposed TFA and PFE remarkably improve the performance of the baseline PhysNet.}
\vspace{-0.8em}
\resizebox{0.49\textwidth}{!} {\begin{tabular}{c|cc|cc|cc}
\toprule[1pt]
\multirow{2}{*}{Method} & \multicolumn{2}{c|}{UBFC} & \multicolumn{2}{c|}{PURE} & \multicolumn{2}{c}{COHFACE}  \\ 
\cline{2-7}
                        & MAE${\downarrow}$    & RMSE ${\downarrow}$     & MAE ${\downarrow}$  & RMSE ${\downarrow}$             & MAE ${\downarrow}$  & RMSE   ${\downarrow}$             \\ 
\hline
CHROM                   & 3.44  & 4.61              & 2.07 & 2.5                & -    & -                      \\
POS                     & 2.44  & 6.61              & 3.14 & 10.57              &  -    &  -                     \\
HR-CNN                  &  -     &  -                 & 1.84 & 2.37               & 8.10  & 10.8                  \\
DeepPhys                & 2.90   & 3.63              & 1.84 & 2.31               &   -   & -                      \\
Zhan et al.             & 2.44  & 3.17              & 1.82 & 2.29               &   -   & -                      \\
Gideon et al.           & 3.60   & 4.60               & 2.30  & 2.90                & 2.30  & 7.60                   \\
AND-rPPG                & 2.67  & 4.07              &  -    &  -                  &   -   & -                      \\
TDM                     & 2.32  & 3.08              & 1.83 & \textbf{2.33}               &    -  &  -                     \\ 
\hline
PhysNet                 & 2.95  & 3.67              & 2.16 & 2.7                &  5.38  &   10.76                 \\
\textbf{TFA-PFE (Ours)}           & \textbf{0.76} & \textbf{1.62}              & \textbf{1.44} & 2.50  & \textbf{1.31} & \textbf{3.92}                  \\
\bottomrule[1pt]
\end{tabular}}
\vspace{-0.8em}
\label{tabel1}
\end{table}

\subsubsection{Results on UBFC-rPPG.} It can be seen from the second column of Table \ref{tabel1} that the vanilla 3DCNN-based PhysNet performs worse than two other deep learning based methods (DeepPhys and TDM). When assembled with the proposed TFA and PFE blocks, the PhysNet+TFA+PFE achieves the best performance on UBFC-rPPG, outperforming the DeepPhys and TDM by -1.86 bpm and -1.28 bpm MAE, respectively. On other words, the proposed TFA and PFE blocks improve the baseline PhysNet performance, and reduce 1.81 bpm MAE and 1.39 bpm RMSE on UBFC-rPPG, indicating the effectiveness of robust rPPG features representation via TFA-PFE blocks.  


\subsubsection{Results on PURE.} As shown in the third column of Table \ref{tabel1}, compared with the baseline PhysNet, the proposed TFA and PFE blocks improve the MAE performance from 2.16 bpm to 1.44 bpm on PURE. It indicates that TFA-PFE is able to represent more motion-robust rPPG features as there are plenty of hard samples with severe head movement in PURE. As for the RMSE metric, the proposed method performs slightly worse than TDM (+0.17 bpm), which is mainly caused by the difference of rPPG backbones (two extra differential temporal convolutions are used in TDM). Please note that the proposed TFA and PFE blocks might also plug-and-play on TDM to further improve performance.


\subsubsection{Results on COHFACE.} Compared with UBFC-rPPG and PURE, the face videos are highly compressed on COHFACE, resulting obvious compression artifacts. As can be seen from the last column of Table \ref{tabel1}, existing supervised CNN-based methods (HR-CNN, Gideon et al.~\cite{gideon_way_2021}, and PhysNet) performs poorly ($\textgreater$5 bpm RMSE) due to the low face video quality. Thanks to the feature refinement from the TFA and PFE blocks, the proposed method achieves state-of-the-art performance (RMSE=3.92 bpm), outperforming previous methods by a large margin. 

\subsubsection{Results on UCLA-rPPG.}
UCLA-rPPG\textbf{} involves more diverse scenarios and presence of subject skin tones. To show our paper is trying to target the general issues of rPPG, we also train and test the TFA and PFE blocks on it. As shown in Table~\ref{ucla}, our method still improve the performance.

\begin{table}[t]
\centering
\caption{The performance of TFA and PFE on UCLA-rPPG}
\vspace{-0.8em}
\resizebox{0.49\textwidth}{!} {\begin{tabular}{c|cc|cc|cc}
\toprule[1pt]
\multirow{2}{*}{Method} & \multicolumn{2}{c|}{Vanilla} & \multicolumn{2}{c|}{PFE} & \multicolumn{2}{c}{PFE+TFA}  \\ 
\cline{2-7}
                        & MAE${\downarrow}$    & RMSE ${\downarrow}$     & MAE ${\downarrow}$  & RMSE ${\downarrow}$             & MAE ${\downarrow}$  & RMSE   ${\downarrow}$             \\ 
\hline
PhysNet                  &11.82 & 17.23  &9.15& 14.82  &8.73& 13.23                     \\
PhysFormer                    &11.75 & 16.39  &8.92& 11.28  &5.96& 12.17                     \\
\bottomrule[1pt]
\end{tabular}}
\vspace{-0.8em}
\label{ucla}
\end{table}

\begin{figure}[t]
\centering
\includegraphics[width=0.8\columnwidth]{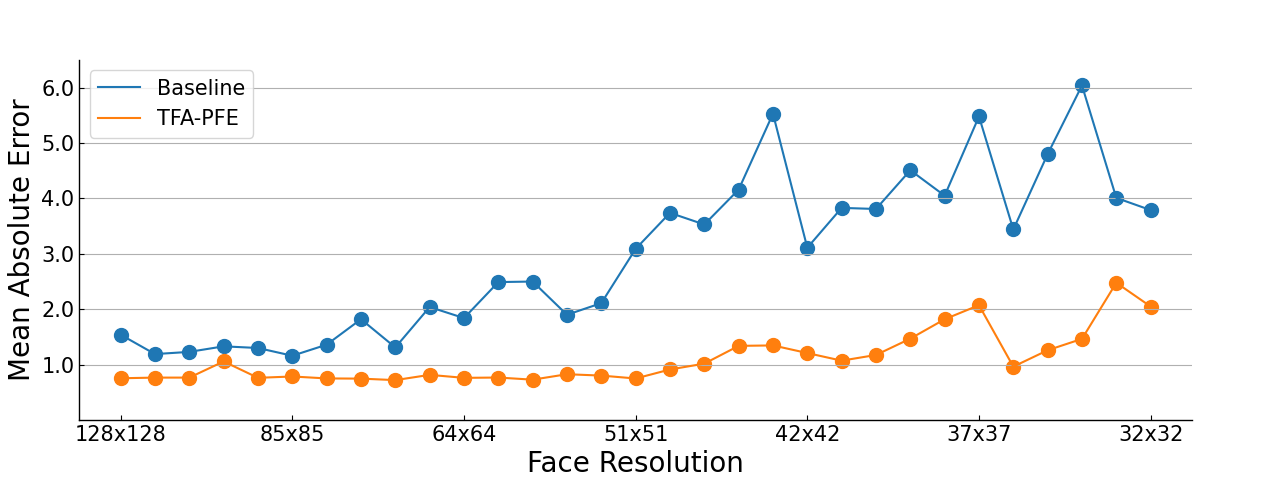} 
\vspace{-0.8em}
\caption{HR estimation results (bpm) on UBFC-rPPG under different fixed face resolution setting.}
\vspace{-1.0em}
\label{fig4}
\end{figure}

\subsection{Comparison on Arbitrary Face Resolution Settings}
Here we downsample the face images from the UBFC dataset with two settings: fixed face resolution and varying face resolution. The former one describes the long-distance scenario while the latter mimics the varying face-camera-distance scenario. 

\subsubsection{Results on fixed face resolution.} It can be seen from the Figure~\ref{fig4} that the baseline PhysNet is easily influenced by the face resolution. When the fixed face resolutions are smaller than 50x50, the performance of PhysNet drops sharply (e.g., MAE$\textgreater$5 bpm). In contrast, when assembled with the proposed TFA and PFE blocks, it can predict accurate rPPG-based HR (MAE$\textless$2 bpm) in most face solution settings.

\begin{table}[t]
\centering
\caption{MAE results (bpm) of the PFE and TFA blocks on UBFC under varying face resolution setting. `128 to 64' means that the face resolution gradually decreases from 128x128 to 64x64 in a video clip.}
\vspace{-0.8em}
\resizebox{1.0\columnwidth}{!}{
\begin{tabular}{|c|c|c|c|c|} 
\hline
\diagbox{Model}{Resolution}   & 128 to 64 & 64 to 32 & 128 to 64 to 128 & 64 to 32 to 64 \\ \hline
Baseline & 10.73  & 8.14  & 11.85      & 7.15     \\
PFE      & 3.49   & 3.84  & 3.29       & 5.59     \\
TFA      & 5.44   & 5.04  & 5.14       & 4.33     \\
\textbf{TFA-PFE}  & \textbf{1.86}   & \textbf{1.97}  & \textbf{1.85}       & \textbf{1.88}     \\ \hline
\end{tabular}
\vspace{-1.5em}
}
\label{tabel3}
\end{table}
\subsubsection{Results on varying face resolution.} rPPG measurement from varying face resolution video is challenging due to the complex temporal contextual interference. The results of varying face resolution on UBFC are shown in Table \ref{tabel3}. Compared with the vanilla PhysNet, the proposed TFA and PFE blocks benefit the facial feature alignment and refinement among consecutive frames, thus improving the MAE performance (8.87 bpm, 6.17 bpm, 10.00 bpm and 5.17 bpm) in four scenarios (high/low resolution gradually decreasing resolution, and high/low resolution gradually decreasing and then increasing resolution), respectively.

\begin{table}[t]
\centering
\caption{MAE results (bpm) of the PFE and TFA blocks on UBFC under different fixed face resolution settings.}
\vspace{-0.8em}
\resizebox{.48\textwidth}{!}{
\begin{tabular}{|c|c|c|c|c|c|} 
\hline
\diagbox{Model}{Resolution}   &  128${\times}$128  & 96${\times}$96 & 85${\times}$85 & 75${\times}$75 & 64${\times}$64 \\ 
\hline
Baseline    & 2.38  & 2.64  & 2.21  & 6.06  & 6.82   \\ 
\hline
PFE w/o RAE         & 3.79  & 3.77 & 3.80  & 3.73  & 3.76  \\
PFE w/o $\mathcal{L}_{crc}$       & 2.87  & 2.85 & 2.89  & 2.86 & 2.84  \\
PFE     &2.40		&2.39	&2.41	&2.34	&2.33\\ 
\hline
TFA   & 8.73  & 8.30  & 8.72 & 8.76  & 8.29  \\
TFA(single)-PFE  & 2.28  & 2.29 & 2.35  & 2.35  & 2.44  \\ 
\textbf{TFA-PFE}    & \textbf{1.60}  & \textbf{1.61} & \textbf{1.62}  & \textbf{1.60} & \textbf{1.63}  \\
\hline
\end{tabular}
}
\vspace{-1.2em}

\label{tabel2}
\end{table}
\subsection{Ablation Study}
We also provide the ablation studies of the PFE and TFA blocks under arbitrary face resolution and severe head movement scenarios on the UBFC dataset. 

\subsubsection{Efficacy of the cross-resolution constraint.}
In the default setting, the models with PFE are trained in two views with different face resolution frames using a cross-resolution constraint $\mathcal{L}_{crc}$. In this ablation, we consider how $\mathcal{L}_{crc}$ impacts the PFE. As shown in Table~\ref{tabel2}, `PFE' outperforms `PFE w/o $\mathcal{L}_{crc}$' by a convincing margin (0.4 to 0.5 bpm MAE) for almost all different face resolution settings. 
It indicates such simple resolution consistency benefits the PFE learn resolution-robust rPPG cues.

\subsubsection{Efficacy of the PFE block on arbitrary face resolution.}
Here we investigate the impacts of the representative area encoding (RAE) of PFE on fixed face resolution UBFC first. It can be seen from the results `PFE w/o RAE' and `PFE' in Table \ref{tabel2} that the PFE without RAE performs even worse than the baseline PhysNet itself under high-resolution cases. When equipped PFE with RAE, it can achieve robust HR estimation under all different fixed face resolution settings. Besides, under more challenging varying face resolution scenarios, we can also find the consistent conclusion from Table~\ref{tabel3} that PFE significantly improves the baseline performance (reducing 7.24 bpm MAE for high resolution gradually decreasing resolution scenario).



%

\subsubsection{Efficacy of the TFA block on arbitrary face resolution.} As shown in the result of `TFA' in Table \ref{tabel2}, the baseline with only TFA block performs even worse than the baseline itself. It is because when estimating the optical flow, all clips are interpolated to a fixed resolution, which makes the TFA block be weak in describing rPPG-aware color area~\cite{xue_video_2019}. We can find from the result `TFA-PFE' that the best performance can be achieved for all different fixed face resolution settings when assembling baseline with both PFE and TFA blocks. Similar conclusion can be also drawn from the varying face resolution setting in Table~\ref{tabel3}. Moreover, we also consider the online testing case that the temporal alignment state from backward frames is not available. In this case, we design a TFA block with a single forward direction for facial feature alignment. It can be seen from the result `TFA(single)-PFE' in Table~\ref{tabel2} that the MAE performance degrades sightly (around 0.6 bpm) compared with bi-directional TFA, but still can improve the performance of the model compared with baseline with only PFE.

\begin{figure}[t]
\centering
\vspace{-1.2em}
\includegraphics[width=0.88\columnwidth]{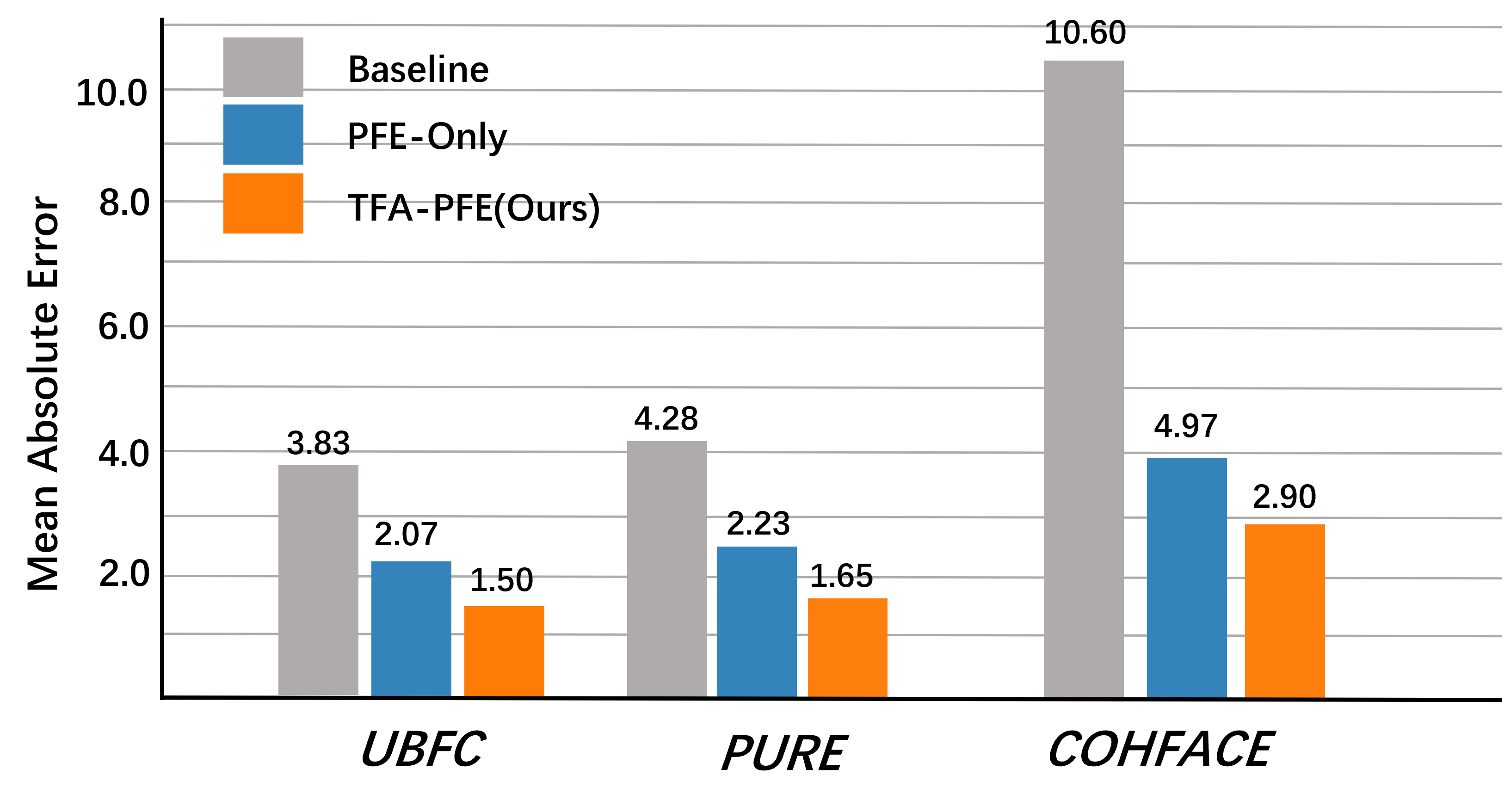} 
\vspace{-1.2em}
\caption{HR estimation results (bpm) on the samples with severe head movement and huge face rotation. }
\vspace{-0.6em}
\label{fig8}
\end{figure}

\subsubsection{Efficacy of the PFE and TFA blocks on severe head movement.} To confirm the efficacy of the PFE and TFA blocks on severe head movements with large angle rotation, we also conduct studies on the carefully selected videos with large angle rotation from UBFC, PURE and COHFACE. Specifically, the participants in COHFACE quickly rotate their heads at an average angle of 80\textdegree, while the participants in UBFC and PURE rotate their heads very slowly at an average angle of 35\textdegree. The results are shown in Figure~\ref{fig8}. We can find that 1) compared with baseline PhysNet, the PFE obviously improves the performance on these videos with  severe head movement, and reduces MAE by 1.76 bpm, 2.05 bpm, and 5.63 bpm on UBFC, PURE and COHFACE, respectively; 2) the proposed TFA-PFE further decreases the MAE by 0.57 bpm, 0.58 bpm, and 2.07 bpm on these datasets due to the excellent motion-robust capacity of TFA.

\begin{table}[t]
\centering
\caption{The number of parameters and FLOPs for the model and inference time on edge devices.}
\vspace{-0.8em}
\resizebox{0.95\columnwidth}{!}{
\begin{tabular}{|c|c|c|c|c|} 
\hline
\textbf{Model}&Parameters&FLOPs&Xavier AGX&Agx orin \\
\hline

Physnet&768.64KB &35.00G&0.18s&0.25s \\
Physnet+PFE &805.77KB &46.51G&2.24s&3.63s \\
Physnet+PFE+TFA &1344.67KB &92.67G&3.37s&4.82s \\
Physformer&7.37MB &50.48G &1.83s&2.18s \\
Physformer+PFE &7.42MB &81.52G &2.47s&3.81s \\
Physformer+PFE+TFA &8.24MB &129.59G &3.52s&4.67s \\
\hline
\end{tabular}
}
\vspace{-1.2em}
\label{Edge}
\end{table}
\subsubsection{Edge Deployment.}
Considering that rPPG will be employed mostly on edge devices with limited computational capacity, we provide a detailed analysis of our model in Table \ref{Edge}.

\vspace{-0.6em}
\section{Conclusion}
In this paper, we propose two plug-and-play blocks, namely TFA and PFE, for remote physiological measurement. With temporal face alignment block and physiological signal structure feature extraction block, the baseline model is able to achieve superior performance on benchmark datasets both on high and low resolution. These blocks are still computationally expensive. Future directions include: 1) to design more lightweight TFA block to escape from optical flow calculation to achieve higher performance; 2) further research on the influence of other video qualities such as arbitrary frame rate and compression rate.

\vspace{0.2em}

\noindent\textbf{Acknowledgment}  \quad This work was supported by the National Natural Science Foundation of China (Grant No. 62002283), and the Fundamental Research Funds for the Central Universities.

\begin{figure*}[t]
\centering
\includegraphics[width=0.85\textwidth]{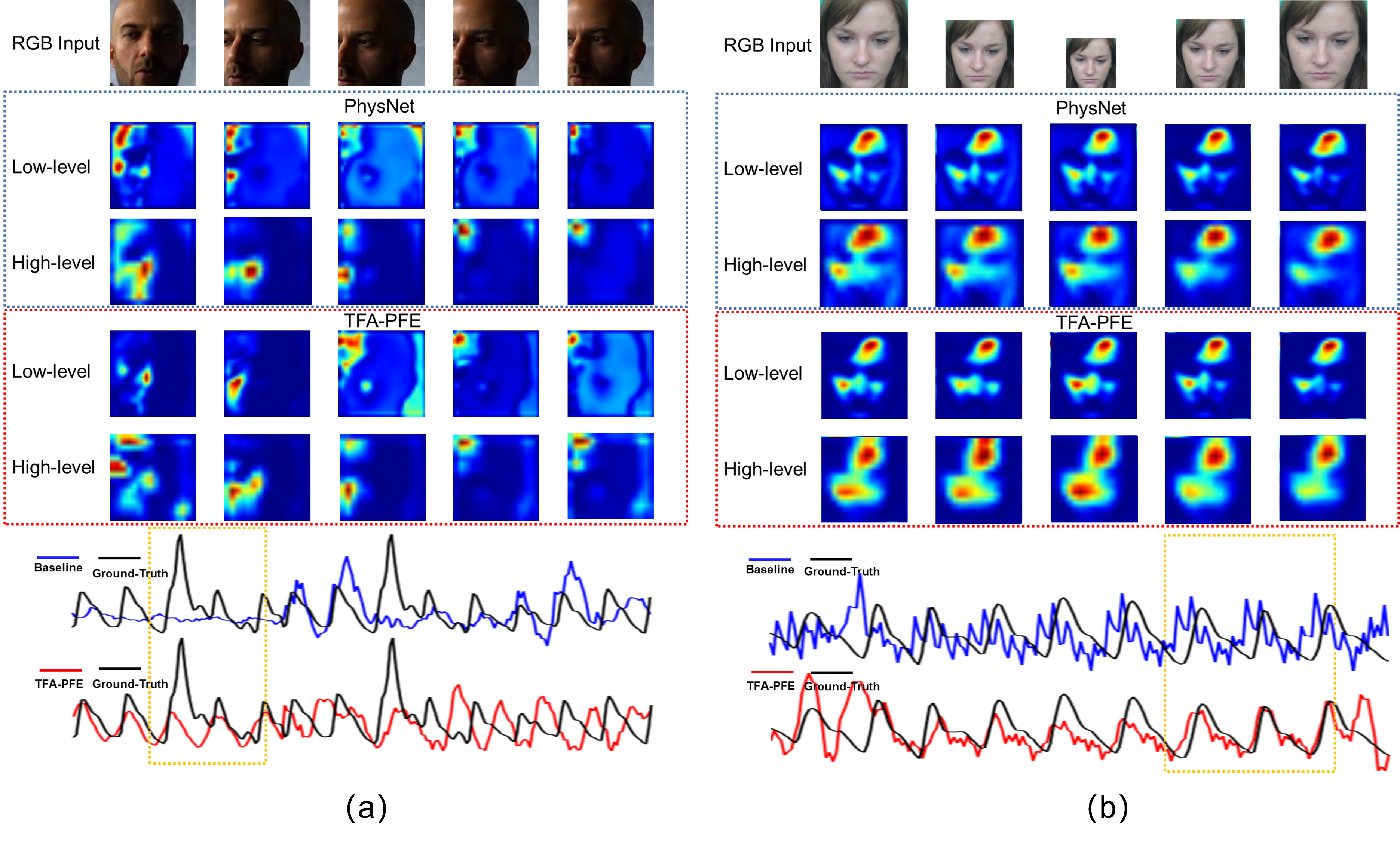}
\vspace{-0.8em}
\caption{Visualization of Baseline (PhysNet) and PhysNet with the TFA-PFE blocks. (a) The feature heatmap on severe head movement. (b) The feature heatmap on varying face resolution.}
\label{fig2}
\end{figure*}

\begin{table*}[t]
\centering
\begin{tabular}{|c|c|c|c|c|c|c|c|c|c|} 
\hline
         \diagbox{Model}{Resolution}   &  128${\times}$128  & 96${\times}$96 & 85${\times}$85 & 75${\times}$75 & 64${\times}$64 & 51${\times}$51 & 42${\times}$42 &36${\times}$36 &32${\times}$32   \\ 
\hline
Baseline & 2.38 & 2.64 & 2.21 & 6.06 & 6.82 & 10.55  & 10.47  & 15.16  & 11.57  \\ 
\hline
PFE-3    & 2.41 & 2.40 & 2.41 & 2.34 & 2.33 & 2.48   & 2.34   & 2.46   & 2.28   \\
PFE-5    & 5.48 & 5.48 & 5.48 & 5.49 & 5.48 & 5.515  & 5.53   & 6.4    & 6.215  \\
PFE-7    & 6.90 & 6.92 & 7.08 & 6.90 & 7.09 & 11.815 & 11.165 & 12.085 & 12.66  \\
\hline
\end{tabular}
\caption{MAE results (bpm) of PFE block with different receptive field expansion on UBFC. ${PFE-n}$ means expanding its receptive field via concatenating the ${n\times n}$ neighboring features. }
\label{table1}
\end{table*}

\begin{table*}[t]
\centering
\begin{tabular}{|c|c|c|c|c|c|c|c|c|} 
\hline
\diagbox{Model}{Resolution} &      &  128${\times}$128  &  64${\times}$64   &  32${\times}$32  & 128 to 64 to 128 & 128 to 64 & 64 to 32 to 64 & 64 to 32  \\ 
\hline
Baseline(PhysFormer) & RMSE & 2.41 & 2.52 & 3.58 & 2.26       & 2.43   & 2.49     & 3.50   \\
                     & MAE  & 1.36 & 1.44 & 1.99 & 1.04       & 1.36   & 1.44     & 1.91   \\ 
\hline
PhysFormer-PFE-TFA   & RMSE & 1.72 & 1.59 & 3.18 & 1.82       & 1.98   & 3.18     & 3.16   \\
                     & MAE  & 0.86 & 0.73 & 1.69 & 0.97       & 1.15   & 1.51     & 1.63   \\
\hline
\end{tabular}
\caption{MAE and RMSE results (bpm) of PhysFormer with TFA and PFE block on UBFC. The results show that it improves the performance of PhysFormer. `128 to 64' means that the face resolution gradually decreases from 128x128 to 64x64 in a video clip.}
\label{table2}
\end{table*}

\section{Appendix A: Feature Visualization}
The low-level features and high-level features of the baseline PhysNet\cite{yu_remote_2019}  and the proposed method (i.e., PhysNet with the proposed TFA and PFE blocks) are visualized in Figure \ref{fig2}. 

As shown in Figure \ref{fig2}(a), in severe head movement (rotation angle exceeds ${45^{\circ}}$) and poor illumination  circumstance (more than half of the face is in the dark), the proposed TFA and PFE blocks could help to better trace the rPPG signals. Specifically, with TFA-PFE, the neural network successfully focuses on the region of interest (ROI) that are not influenced by pose and illumination variation(e.g., the ROIs on heatmap are larger and more reasonable than the baseline PhysNet).

Figure \ref{fig2}(b) shows that, when dealing with the problem of varying face resolution, the TFA and PFE blocks could consistently focus on the significant ROIs to extract rPPG signals. The ROIs with bright color are all distributed on the regions with abundant blood vessels.

\section{Appendix B: Receptive Field ${n}$ of PFE Block}
In this section, we discuss about the selection of parameter $n$ in PFE block for receptive field expansion. As shown in Table \ref{table1}, it achieves the best performance when $n$ is selected as $n=3$. We can conclude that the receptive field expansion averages the subtle change of rPPG on the local pixels and reduces the influence of noises. But if the expansion ratio is too large (i.e., with a larger $n$), the performance will be decreased since multiple disturbance could be mixed together in that case.\cite{lee2020meta}
Moreover, the channel number of the feature maps is ${n^{2}C}$.  Thus, it could also lighten the computation burden with ${n=3}$.

\section{Appendix C: Efficacy of PFE and TFA Blocks on PhysFormer}
To verify the architecture generalization capacity of the proposed PFE and TFA blocks, we also assemble our blocks in the latest state-of-the-art rPPG signal backbone PhysFormer \cite{yu2022physformer}. The results can be seen in Table \ref{table2}. The PFE and TFA blocks improve the performance of vanilla PhysFormer in all face resolution settings. It demonstrates the effectiveness and robustness of the proposed method.

\bibliography{through}

\end{document}